\documentclass[10pt,letterpaper]{article}

\usepackage{cogsci}

\cogscifinalcopy 

\usepackage{pslatex}
\usepackage{apacite}
\usepackage{float} 

\title{Analysing the Impact of Audio Quality on the Use of Naturalistic Long-Form Recordings for Infant-Directed Speech Research}

\author{{\large \bf Mar\'ia Andrea Cruz Bland\'on (maria.cruzblandon@tuni.fi)} \\ Unit of Computing Sciences, Tampere University, P.O. Box 553, FI-33101, Finland \AND {\large \bf Alejandrina Cristia (alecristia@gmail.com)} \\ Département d’études cognitives, ENS, EHESS, CNRS, PSL University, 29 rue d'Ulm, 75005 Paris, France \AND {\large \bf Okko R\"as\"anen (okko.rasanen@tuni.fi)}\\ Unit of Computing Sciences, Tampere University, P.O. Box 553, FI-33101, Finland}
  
\usepackage[table]{xcolor} 

\usepackage{graphicx}
\usepackage{subfig} 
\usepackage{multirow}

\newcommand{\STAB}[1]{\begin{tabular}{@{}c@{}}#1\end{tabular}}

\begin{document}

\maketitle

\begin{abstract}

Modelling of early language acquisition aims to understand how infants bootstrap their language skills. The modelling encompasses properties of the input data used for training the models, the cognitive hypotheses and their algorithmic implementations being tested, and the evaluation methodologies to compare models to human data. Recent developments have enabled the use of more naturalistic training data for computational models. This also motivates development of more naturalistic tests of model behaviour. A crucial step towards such an aim is to develop representative speech datasets consisting of speech heard by infants in their natural environments. However, a major drawback of such recordings is that they are typically noisy, and it is currently unclear how the sound quality could affect analyses and modelling experiments conducted on such data. In this paper, we explore this aspect for the case of infant-directed speech (IDS) and adult-directed speech (ADS) analysis. First, we manually and automatically annotated audio quality of utterances extracted from two corpora of child-centred long-form recordings (in English and French). We then compared acoustic features of IDS and ADS in an in-lab dataset and across different audio quality subsets of naturalistic data. Finally, we assessed how the audio quality and recording environment may change the conclusions of a modelling analysis using a recent self-supervised learning model. Our results show that the use of modest and high audio quality naturalistic speech data result in largely similar conclusions on IDS and ADS in terms of acoustic analyses and modelling experiments. We also found that an automatic sound quality assessment tool can be used to screen out useful parts of long-form recordings for a closer analysis with comparable results to that of manual quality annotation.  

\textbf{Keywords:} 
infant-directed speech; benchmarking; long-form recordings; acoustic analyses; computational modelling; early language acquisition
\end{abstract}

\section{Introduction}

Computational modelling of early language acquisition aims to understand how infants bootstrap their language skills. Development of plausible models and using them for cognitive hypothesis testing requires attention on three largely independent aspects: models themselves, the language data used as inputs for the models, and evaluation protocols for assessing linguistic competence or performance of the models.

Given the advances in machine learning and data collection methods, computational models have become increasingly capable of learning from semi-naturalistic large-scale auditory \cite{Lavechin2022} or audiovisual data \cite{Khorrami2021}. This is in contrast to earlier studies with simplified inputs, such as phonetic transcripts \cite{Brent1999}, pre-segmented units of speech  \cite<e.g.,>{deBoerKuhl2003}, and small-scale speech corpora often associated with such studies. This development has been largely facilitated by the research on so-called self-supervised learning (SSL; \citeNP<e.g.,>{Oord2018}). These methods enable autonomous speech representation discovery from low-level sensory inputs without strong linguistic priors to guide the learning process. Therefore, they are also highly relevant also as potential models of language learning. As a result, recent research has shown successes in modelling phonemic and lexical learning from acoustic speech only \cite{Lavechin2022}, including also models learning from naturalistic child-centred long-form audio recordings \cite{Lavechin2022daylong}. 


In terms of model evaluation, there is also ongoing progress in working towards more realistic model evaluation protocols with respect to experimental or observational findings on infant language development. For instance, earlier studies that have compared models against individual behavioural experiments \cite{Nixon2021PredictionModel} or analysed developmental patterns \cite{Lavechin2022, nikolaus-etal-2022-learning}. \citeA{Cruz2021} also recently proposed an evaluation framework where models are subjected to computational tests comparable to those used with real infants in lab, allowing comparison of model language capabilities to those of infants as a function of model training data and infant age. 

Notably, behavioural studies and hence their computational replications typically employ stimuli that have been carefully curated so that researchers can control the dimensions of the stimulus that are of interest for the study. This, in turn, can potentially lead to less authentic and simpler stimuli to what infants are actually exposed to in real world. There is evidence that using in-lab data might overestimate the effect of linguistic capabilities when contrasted with more natural/spontaneous data \cite{MacDonald2020}. Therefore, a logical step would be to extend computational model evaluation towards more naturalistic data, such as use of audio stimuli extracted from child-centred long-form recordings from children's everyday environments \cite<e.g.,>{Bergelson2019}. 

However, one central problem with using naturalistic recordings is their highly variable and on average very low audio quality compared to cleaner corpora recorded in controlled settings (e.g., almost 0 dB average signal-no-noise ratio reported by \citeNP{RASANEN_2019}). The noise in audio may influence results of acoustic analyses and computational modelling studies, and also introduces a potential mismatch between datasets used to train and evaluate models. Hence, it would be important to separate the effects of moving from simplified to naturalistic input from those of moving from clean data to potentially more noisy data. While the former is relevant for motivating the use of more naturalistic data, the latter is simply a potential nuisance factor in analyses, unless the utilised acoustical analyses and computational models are specifically equipped against noise (which is rarely the case).

The present paper uses analysis of infant-directed speech (IDS) and adult-directed speech (ADS) as a case study to address the above issues. We investigate whether 1) basic properties of naturalistic IDS and ADS speech data differ from those of in-lab stimuli, and 2) investigate whether computational modelling findings with naturalistic are different from in-lab data. Importantly, we perform both investigations with different sound quality screening methods applied to the data in order to test how varying sound quality criteria impact study findings, and in order to identify the best practice for creating high-quality stimulus sets from longform audio recordings. As our computational model, we employ a self-supervised algorithm for statistical learning investigated by \citeA{Lavechin2022daylong}.

\section{Methodology}
The aim of the study was to create a naturalistic test set of IDS and ADS utterances that can be used to benchmark computational models of language acquisition in their capability to process IDS and ADS differentially, and to study its properties compared to in-lab data as a function of different sound quality criteria.

We chose to study IDS due to its pivotal role in language bootstrapping, guiding infants' attention to the relevant parts of utterances and making contrasts more easy to identify \cite{SODERSTROM_2007, Golinkoff_IDS_2015}. IDS has been found to be widely used across different languages and communities, and to display common features that varies in strength depending on the language \cite{Cox2023}. Typically IDS will have a more salient pitch contour, longer vowels and simpler grammar than ADS \cite{SODERSTROM_2007, hilton2022acoustic}. In addition, there exists carefully curated in-lab data with well-established research findings for IDS and ADS comparison from the ManyBabies (MB) consortium \cite{Frank2020QuantifyingPreference} as well as long-form recordings annotated for speech addressee \cite{Bergelson2019, Canault2016}. Finally, there are earlier modelling findings using both in-lab and naturalistic IDS data, where the authors reported smaller effects for naturalistic data compared to the in-lab data \cite{MacDonald2020}. However, the authors did not control for sound quality in their experiments, which could potentially explain the difference in their findings between the two data types.

What then makes long-form child-centred recordings typically noisier than their in-lab counterparts is that they include background noises (e.g., radio or television noise), overlapped speech, baby cries and screams among other noises in addition to speech varying from near- to far-field recordings. Hence, long-form recordings are more challenging to perceive and process for annotators and computational models alike. Therefore, in this study, we extend the comparison of long-form recordings against in-lab data by introducing the audio quality as a tentative factor moderating the results. Additionally, we run the analyses for two rhythmically different languages: English and French.  Finally, we also evaluate the impact of the audio quality in model evaluation by using the developed test set with self-supervised models for predictability of IDS and ADS utterances (see,  \citeNP<e.g.,>{MacDonald2020,RASANEN_2018} for earlier related modelling approaches). 

\subsection{Corpora for Benchmarking IDS and ADS Segments from Long-Form Recordings}
\label{sec:corpora}
For creation of a naturalistic test set of IDS and ADS stimuli, we employed the hand-annotated IDS and ADS segments of the English part of the ACLEW project data \cite{Bergelson2019, soderstrom2021developing} (hereafter, SEEDLings data) and the annotated segments of the Lyon project data \cite{Canault2016} for the French dataset (hereafter, Lyon data). Both SEEDLings and Lyon data were collected by placing a LENA audio recorder device in the clothes of the infant and recording a whole day. The annotated segments were obtained by randomly sampling two-minute (SEEDLings) or ten-minute (Lyon) segments from a full recording of an infant, and for ten infants ($3-20$ months of age) in the SEEDLings data and 16 infants ($3-41$ months) in the Lyon data. Authors of these datasets had the clips manually annotated for utterances\footnote{In contrast to \citeA{MacDonald2020} who used LENA-based automatic processing followed by manual verification of labels.} and their speakers and addressees.

For our experiments, we used the utterances from adult speakers (female or male) with adult addressee as ADS and utterances from adult speakers with infant addressee as IDS.  This yielded a total of 1124 ADS utterances and 952 IDS utterances for the SEEDLings data, and 3915 ADS utterances and 863 IDS utterances for the Lyon data. These utterances were further subjected to sound quality assessment, as explained in the next subsection. 

In order to compare the utterances from the long-form recordings against stimuli used in behavioural experiments, we chose the MB project stimulus set as the reference, consisting of 120 IDS and 120 ADS utterances \cite{Frank2020QuantifyingPreference}. The MB recordings were carried out using a LENA device or a mobile device, but conducted in a fixed lab environment with an elicitation task related to object description to an infant or adult (see \citeNP{Frank2020QuantifyingPreference}, for details).

\subsection{Assessing the Audio Quality of IDS and ADS Segments}
\label{sec:quality}
To determine the audio quality of the naturalistic IDS and ADS segments, we used two strategies: manual and automatic annotation. For the manual strategy,  three labels were utilised: good, medium and bad quality. The quality was defined in terms of perceived background noise, overlapping speech, baby cries and screams, and loudness and reverberation of the target speech. The distinction between good and medium quality was based on the loudness and distance to the microphone (i.e., good quality required high loudness and little reverberation). One annotator (the first author) annotated all the samples in both datasets across several days. 

Automatic sound quality ratings were obtained from a recent Signal-to-Noise Ratio (SNR) estimator system \cite{Brouhaha2022}. SNRs (in dBs) of all utterances were first estimated. The manual annotations were then used to establish the SNR threshold that best divides the data into good and medium/poor quality data. This procedure yielded thresholds of 14.50 dB for SEEDLings and 14.03 dB for Lyon data. 

\begin{figure}
    \centering
    \includegraphics[width=0.50\textwidth]{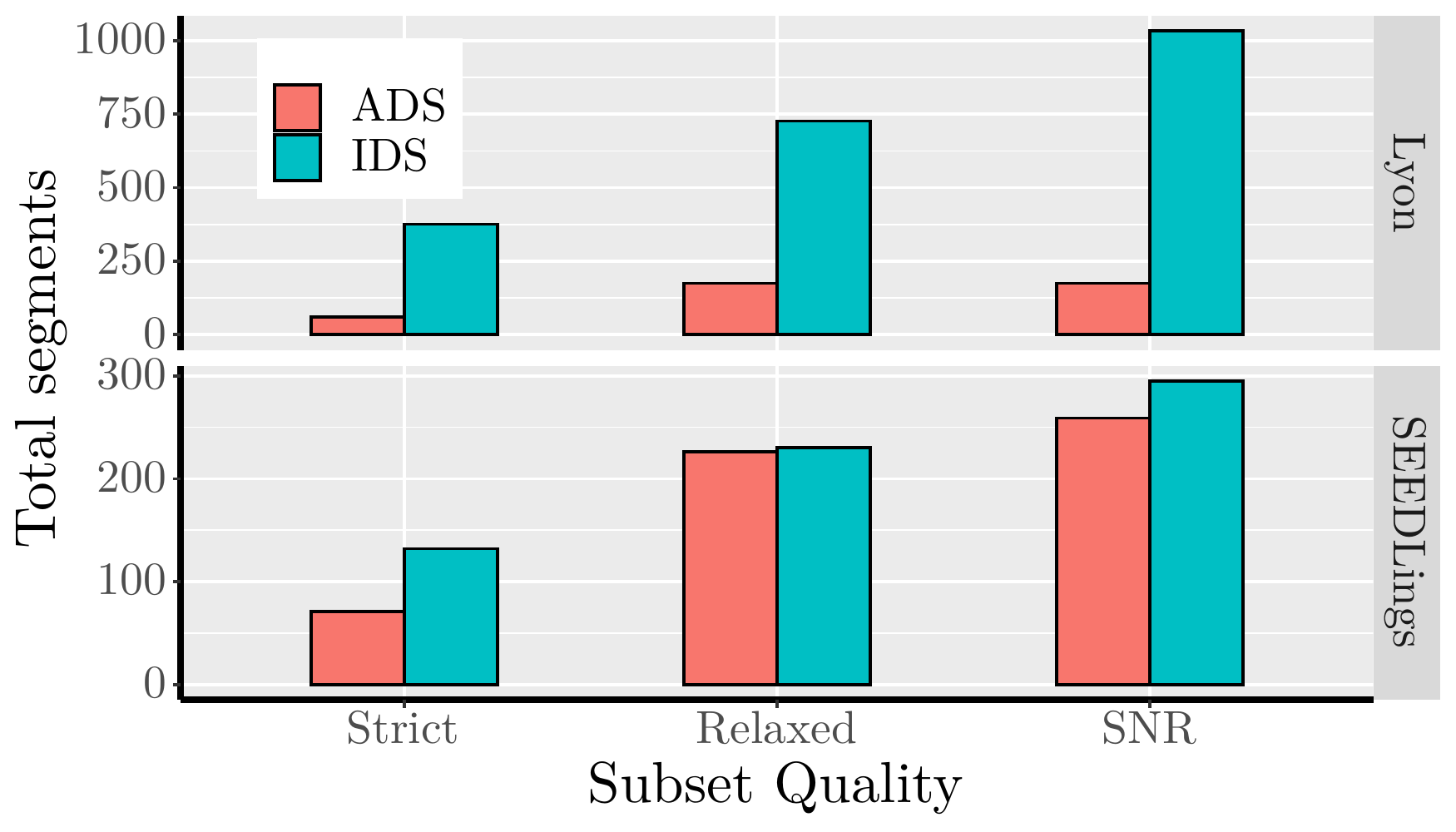}
    \caption{Distribution of IDS and ADS segments according to the determined quality of the segments for the Lyon (top) and SEEDLings (bottom) datasets.}
    \label{fig:distribution}
\end{figure}

Given the ratings, we defined three subsets of data for our experimental analyses:  \textit{Strict quality}, consisting of manual good quality utterances. \textit{Relaxed quality}, consisting of utterances with good and medium quality manual labels, and \textit{SNR quality} utterances whose automatic estimates were above the set SNR thresholds. Figure~\ref{fig:distribution} shows the resulting sample counts of IDS and ADS utterances in the different quality subsets. As seen from the figure, there is a tradeoff between sound quality and test set size with SNR approach resulting in least strict quality requirements with the largest sample size.

\subsection{Acoustic Analyses of IDS and ADS Characteristics}

Following \citeA{MacDonald2020}, we performed acoustic analyses to compare long-form recordings and in-lab recordings in terms of their basic IDS/ADS characteristics.  For this, \textit{the mean of logarithm of fundamental frequency} (log-F0) and \textit{log-F0 variability} (std log-F0) were extracted for all utterances with PRAAT\footnote{By means of the python library wrapper: Parselmouth \cite{parselmouth}}. Additionally, the speech delivered to infants tend to be shorter than the speech delivered to other adults \cite{SODERSTROM_2007}. Thus, we included \textit{utterance duration} (in seconds) in our analyses. Finally, there is evidence that speakers use an exaggerated articulation for IDS \cite{Cox2023} to help the processing of the message. Although extraction of vowel space measures from data without phonetic annotation is not possible, we used  \textit{mean spectral tilt} as a proxy for general vocal effort that also reflects potential differences in speaking style. The first Mel-frequency cepstral coefficient (MFCC) coefficient was used for this purpose.

Overall, we expected to replicate \citeA{MacDonald2020} findings for both languages, i.e., higher pitch and variability for IDS than ADS utterances. For duration,  we expected the duration to be longer for in-lab recordings, and to be independent of the subset quality on naturalistic data. For tilt, we expected greater tilt for utterances with greater vocal effort. Unpaired t-tests were used to identify if there were any significant differences between IDS and ADS acoustic features, and we also report the magnitude of the difference using Cohen's \textit{d} \cite{Lipsey2001PracticalMeta-analysis}. 

\subsection{Computational Modelling Analyses}
\label{sec:modelling}

The aim of the modelling experiments was to test if 1) experiments on naturalistic data replicate behaviour observed on in-lab data, and 2) whether this is moderated by quality screening applied to the data. While the former is not necessarily expected (the reason to use naturalistic data in the first place), it is important to know if models' behaviours are greatly affected by sound quality of the test data.  
\subsubsection{Self-Supervised Model for Speech Representation Learning}
We used the self-supervised statistical learning neural network model introduced in \cite{Lavechin2022daylong} as a model of early language acquisition. The model learns directly from the raw speech signal in an unsupervised manner. The learning process is centred on optimisation of a predictive process that attempts to predict future speech observations from current speech input. However, instead of predicting the acoustic signal directly, the model learns to predict its own latent representations of the future speech, allowing it to also invent these representations as seen fit for the prediction task (cf. brains predicting their own neural activations for sensory input instead of the actual physical signals). The reader is recommended to see \citeA{Oord2018} for technical details of the learning mechanism. 

The model was trained on read audiobooks from LibriVox and LibriLight, training a model for English and French \cite{LibriVox, Librilight} separately. We used the model versions from \citeA{Lavechin2022daylong} with a total training of 50, 100, 200, 400, 800, 1600 or 3200 h to simulate different amounts of language experience for the learner.

\subsubsection{Predictability Analysis}
Predictability of speech is linked to the encoding of speech structure. Models that are better at predicting the input exhibit better comprehension of the speech signal, whereas models that find the input difficult to predict can be viewed as more "surprised" with the stimuli. Previous work has linked lower predictability of IDS (compared to ADS) to heightened attentional preference towards IDS (\citeNP{RASANEN_2018,MacDonald2020}; see also \citeNP{Cruz2021}). However, here we do not take a position in the familiarity/novelty coding of attention, but simply use predictability of input as a measure of model's competence with the given speech style. This allows us then to compare potential mismatches between datasets in terms of naturalness and noise levels.

In the case of these models trained on audiobooks, they have not seen examples of IDS or ADS styles. Nevertheless, based on the previous findings \cite{Lavechin2022daylong}, we expect the models to have better speech representations and higher prediction accuracy as they become trained with more data. If the models are learning speech representations that are encompassing features prevalent across the different speech styles, we expect to have similar predictability for both IDS and ADS. Another option is to have better predictability for ADS than IDS utterances as ADS is more similar to the read speech the models have seen during training. If the quality of the test subset does not affect the behaviour, we expect to have the same predictability pattern across all three qualities. We expect that predictability findings between in-lab and long-form data might differ. This is since the models are focusing on the entire acoustic signals instead of a small set of chosen characteristics, and hence data with acoustically similar measures may still have notable structural differences from the perspective of speech perception.

\section{Results}

\subsection{Acoustic Analyses}
\begin{table*}[t]
\centering
\caption{Results of the acoustic analysis: mean pitch (F0), and variability of the pitch in log scales, spectral tilt, and duration (s) for IDS and ADS segments are reported for all three audio quality subsets. Effect size is reported as Cohen's d, asterisk indicating all significant differences (at  $\alpha=0.05$). Bold font denotes the larger values in cases of significant difference between the IDS and ADS.}
\label{tab:acoustic_features}
\begin{tabular}{clllllllllllll}
\hline
\multicolumn{2}{c}{\multirow{2}{*}{}} & \multicolumn{3}{c}{\textbf{mean(log-F0)}} & \multicolumn{3}{c}{\textbf{std(log-F0)}} & \multicolumn{3}{c}{\textbf{Spectral Tilt}} & \multicolumn{3}{c}{\textbf{Duration (s)}} \\
\multicolumn{2}{c}{} & \multicolumn{1}{c}{IDS} & \multicolumn{1}{c}{ADS} & \multicolumn{1}{c}{d} & \multicolumn{1}{c}{IDS} & \multicolumn{1}{c}{ADS} & \multicolumn{1}{c}{d} & \multicolumn{1}{c}{IDS} & \multicolumn{1}{c}{ADS} & \multicolumn{1}{c}{d} & \multicolumn{1}{c}{IDS} & \multicolumn{1}{c}{ADS} & \multicolumn{1}{c}{d} \\ \hline \\
\multirow{3}{*}{\STAB{\rotatebox[origin=c]{90}{SEEDLings}}} & Strict & \textbf{5.31} & 5.07 & 0.80* & 0.22 & 0.20 & 0.16 & 135.71 & 137.23 & 0.04 & 1.22 & 1.46 & 0.27 \\[1.2pt]
 & Relaxed & \textbf{5.32} & 5.10 & 0.77* & 0.23 & 0.22 & 0.07 & 133.67 & 133.75 & 0 & 1.37 & \textbf{1.86} & 0.43* \\[1.2pt]
 & SNR & \textbf{5.39} & 5.16 & 0.81* & 0.23 & 0.23 & 0.03 & 135.72 & 135.09 & 0.02 & 1.12 & 1.20 & 0.10 \\[1.2pt] \\
\multirow{3}{*}{\STAB{\rotatebox[origin=c]{90}{Lyon}}} & Strict & 5.46 & \textbf{5.58} & 0.43* & 0.25 & 0.22 & 0.20 & 82.42 & 85.09 & 0.07 & 2.25 & 1.99 & 0.21 \\[1.2pt]
 & Relaxed & 5.45 & \textbf{5.50} & 0.18* & 0.25 & 0.26 & 0.05 & 81.98 & \textbf{90.81} & 0.23* & \textbf{2.51} & 2.24 & 0.19* \\[1.2pt]
 & SNR & 5.46 & \textbf{5.56} & 0.35* & 0.25 & 0.24 & 0.06 & 84.91 & \textbf{91.52} & 0.17* & \textbf{2.15} & 1.86 & 0.24*\\[1.2pt] \\
\multicolumn{2}{c}{ManyBabies} & \textbf{5.39} & 5.19 & 1.18* & 0.28 & 0.29 & 0.03 & \textbf{180.16} & 169.08 & 0.31* & 2.97 & \textbf{5.14} & 1.00* \\ \hline
\end{tabular}
\end{table*}

Table~\ref{tab:acoustic_features} shows the measurements of acoustic features obtained for the two speech styles, two corpora, sound quality subsets, and with the MB data as a reference. The MB data follow the predictions of the literature for three acoustic features (see also \citeNP{MacDonald2020}), where IDS utterances exhibit higher F0, shorter duration, and higher spectral tilt than what is in ADS. 

As for the in-lab versus naturalistic data, the effect sizes (Cohen's d) of acoustic feature differences are in general larger for the MB data than for the long-form recordings, also replicating the findings of \citeA{MacDonald2020}. This adds to the evidence that in-lab recordings tend to be more distinctive between IDS and ADS  than at-home recordings. However, when we look at the patterns found for the SEEDLings and Lyon data, only the SEEDLings dataset replicates the majority of findings of the MB data, except for the spectral tilt, where no significant difference is found between IDS and ADS, and duration, for which only the Relaxed quality replicates the pattern. In contrast to the English data, the French data (Lyon) exhibits a higher mean F0 and tilt for the ADS utterances and longer utterances for IDS than ADS. 

The results for the Lyon corpus contrast with previous findings where the pitch was found higher for IDS than ADS \cite{Cox2023}. During the manual inspection of the audio quality, we found that some of the utterances were annotated as ADS when the adult was in fact talking to another child in the room. This inconsistency in the given annotations could contribute general accuracy of the measurements. However, it cannot explain the reversal of the acoustic cue values in IDS and ADS. Therefore, the reason for the result remains currently unclear, and exemplifies how the analysis of naturalistic data may result in findings that are not expected based on analyses conducted on more controlled datasets.

Regarding the audio quality of the subsets, we observed that in both naturalistic corpora, the measurements of the same acoustic feature slightly vary across the three audio qualities. Generally, the effect sizes are larger for cleaner data (see Cohen's d in Table~\ref{tab:acoustic_features}). However, main effects and their directions observed for the strict audio quality are likely to be found in the less stringent quality screening cases. Hence, from the acoustic analysis point of view, the quality of the recordings does not drastically change the characteristics studied. Since there are no remarkable differences in the results between the manual and automatic quality screening, the findings also suggest that automatic screening is applicable to long-form audio data for basic acoustic analyses. We further investigate whether this finding also holds in the predictability analysis with a computational model.

\subsection{Modelling Analyses}

\begin{figure}
    \centering
    \includegraphics[width=0.50\textwidth]{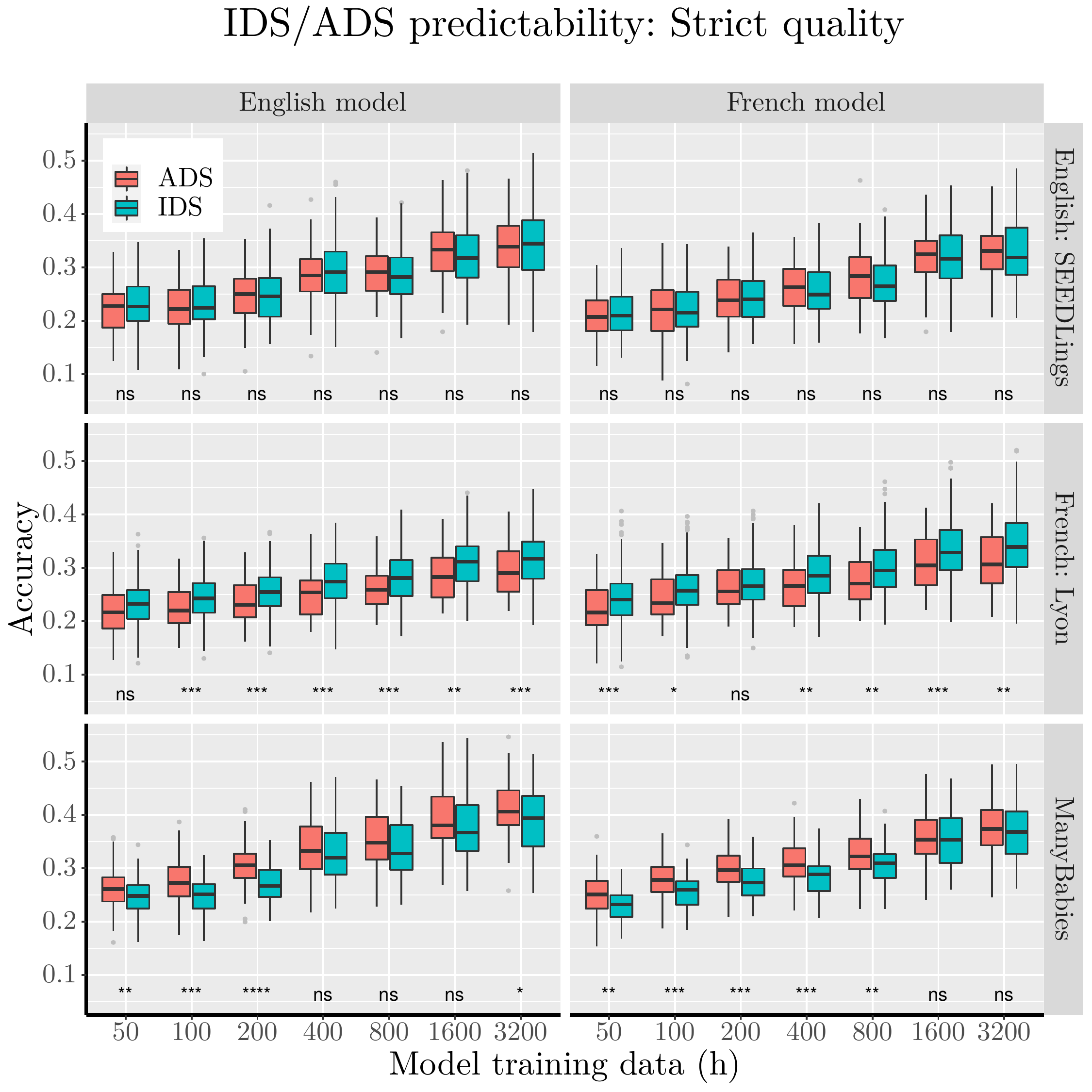}
    \caption{Accuracy of predictions on strict audio quality data for the different model training data sizes (x-axis). Top: SEEDLings data. Middle: Lyon data Bottom: MB data. Left: English-trained model, Right: French-trained model. ADS in red and ADS in green. Significant differences between IDS and ADS are denoted with asterisks or \texttt{ns} for no difference ($*$ stands for $p < 0.05$, $**$ for $p< 0.01$, and $***$ for $p< 0.001$).}
    \label{fig:predictability}
\end{figure}

\begin{figure}
    \centering
    \includegraphics[width=0.50\textwidth]{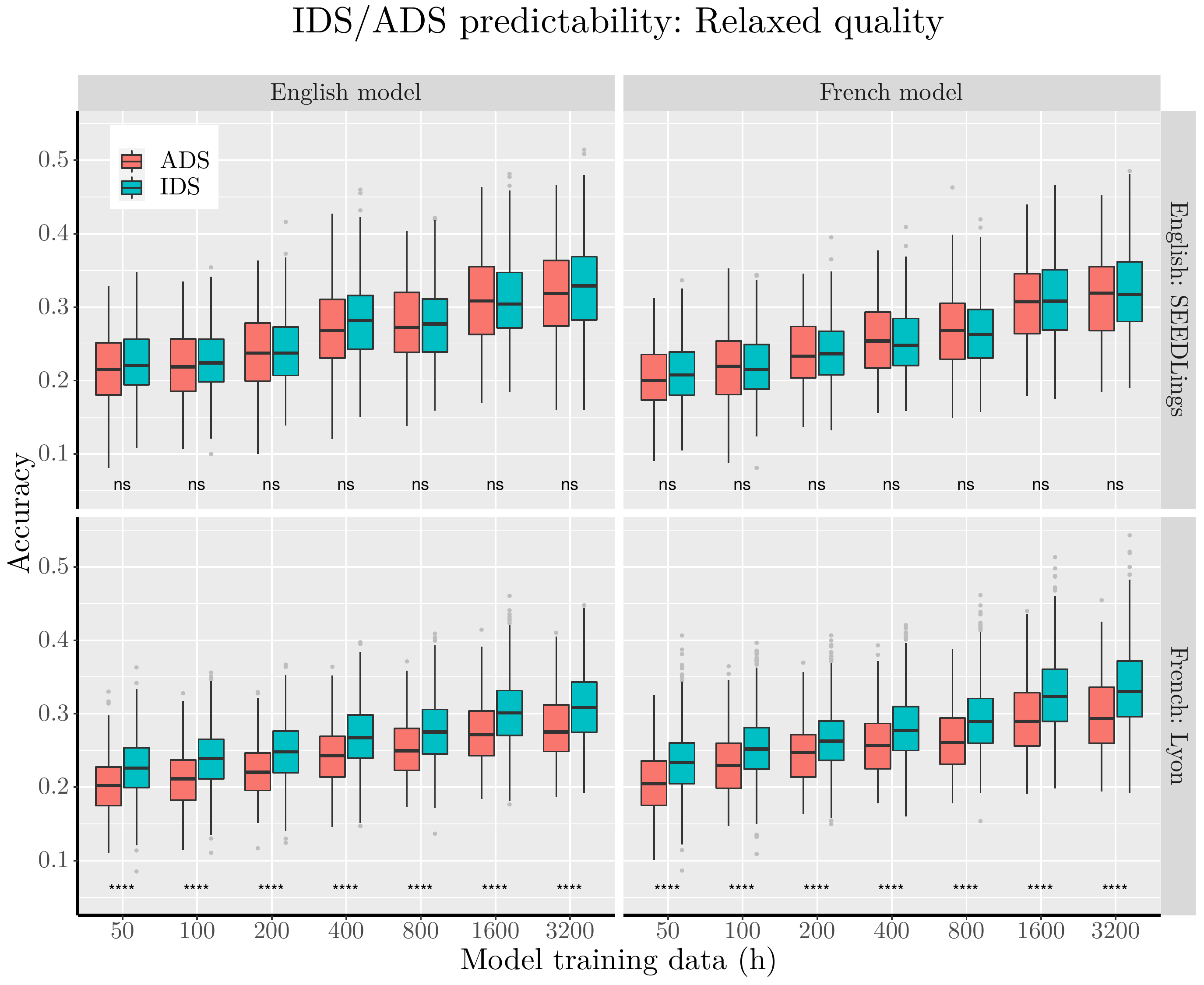}
    \caption{Predictability of IDS and ADS utterances in the relaxed quality subset. See the caption of Fig.~\ref{fig:predictability} for details of the plot.}
    \label{fig:predictability_relaxed}
\end{figure}

\begin{figure}
    \centering
    \includegraphics[width=0.50\textwidth]{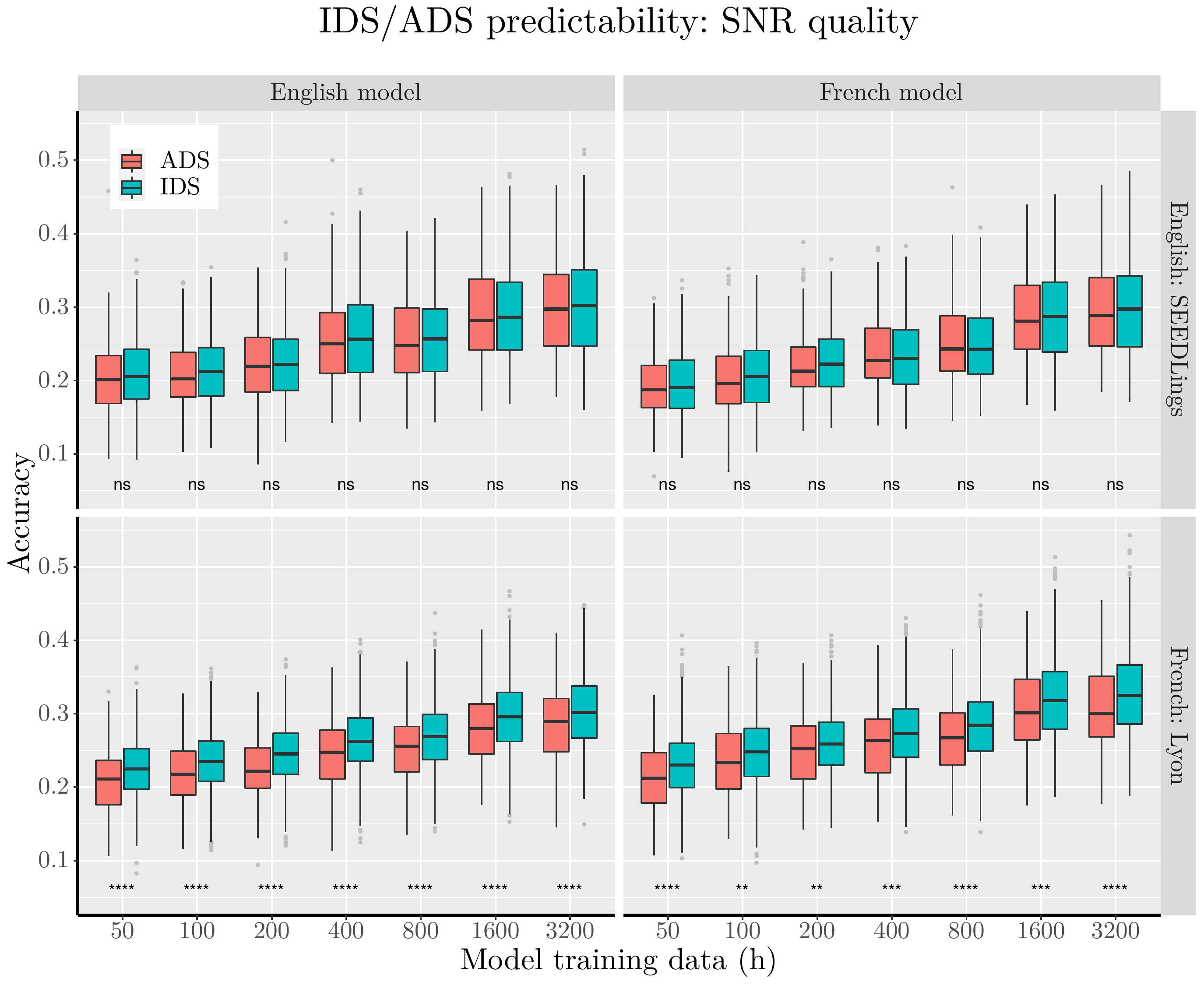}
    \caption{Predictability of IDS and ADS utterances in the SNR quality subset. See the caption of Fig.~\ref{fig:predictability} for details of the plot.}
    \label{fig:predictability_snr}
\end{figure}

Figure~\ref{fig:predictability} presents the prediction accuracy of the self-supervised models for the SEEDLings, Lyon and MB data for the \textit{strict quality}. The results are shown for each of the 7 training data amounts (checkpoints) separately. Figures~\ref{fig:predictability_relaxed} and \ref{fig:predictability_snr} show the results for the \textit{relaxed} and \textit{SNR} qualities for the SEEDLings and Lyon data. 

The results on the in-lab MB data indicate that the model reaches above-chance level accuracy ($> 0.02$) already at 50 h of training, and the predictability improves as the models use more training data; this is regardless of the language used for training. Also, the model exhibits higher predictability of the ADS utterances than those of IDS. However, this difference between the two styles on in-lab data becomes smaller with more training. 

When comparing MB data against the long-form recordings in the strict quality subset (Fig.~\ref{fig:predictability}), we find that only the increasing accuracy pattern is replicated by the SEEDLings and Lyon data. In general, the predictability improves with more training data, again despite of the language of training. However, for the SEEDLings data, there is no significant difference between IDS and ADS. For the Lyon data, the pattern is the opposite from MB, the model obtaining higher predictability for the IDS compared to ADS utterances. These results demonstrate that modelling outcomes on in-lab data and long-form data can differ, as we further discuss in the final section. 

When analysing the results across the three audio quality (see Figs.~\ref{fig:predictability_relaxed} and \ref{fig:predictability_snr}), we can observe that the predictability patterns found for each corpus apply across all the three qualities. As with the acoustic analyses, the predictability becomes slightly worse with more noisy data (e.g., going from accuracy of $\approx0.24$ for the 50 h English model for the SEEDLings data in the strict subset to $\approx0.22$ for the relaxed subset to $\approx0.20$ for the SNR subset). However, the basic pattern of the relative predictability of IDS and ADS as a function of training data amount is not affected by the sound quality.  

Overall, despite being trained on audiobooks, the model is capable of predicting both speech styles with a comparable accuracy, showing no notable difference predicting either style (differences between IDS and ADS do not go greater than $0.1$). The result implies that the model is learning speech representations that are tuned to the properties of spoken language, but the model does not appear to treat IDS/ADS styles in a particularly distinct manner. This happens independently of whether the model is trained on English or French language, although with slightly lower accuracy for the non-native predictions. This may be due to models were not trained on natural ADS or IDS, but on read audiobooks, which may cause a comparable mismatch in speech properties when moving from the read speech to naturalistic speech, no matter whether the naturalistic speech is IDS or ADS.

\section{Discussion and Conclusion}

In order to develop more naturalistic evaluation settings for computational modelling, it is necessary to understand the differences between in-lab and at-home recordings, as well as the implications of  sound quality differences in different datasets. This paper investigated sound quality for the case of IDS/ADS comparison by extracting three audio quality for long-form recordings and contrasting their acoustic features against the features of curated recordings for lab testing. We also used predictability analysis to investigate whether the audio quality could affect modelling findings. 

Overall, there are several differences between in-lab data and the long-form data: contents and duration of the recording (short controlled sessions vs. uncontrolled full-day everyday activities),  social contexts in which the utterances were spoken (see also \citeNP{MacDonald2020}), and with in-lab data displaying more larger acoustic differences between IDS and ADS than those found in long-form data. This suggests that the input infants face in natural contexts may consist of less salient features in comparison to those encountered in-lab contexts. As a consequence, all the differences between the speech styles found in in-lab data might not replicate in at-home data, both in terms of acoustic analysis and computational modelling results (e.g., lack of difference between the spectral tilt of IDS versus ADS). And even in the case of having similar acoustic patterns (e.g., pitch and variability of the pitch for the MB and SEEDLings data), in-lab and at-home data could lead to different conclusions when evaluating computational models (see e.g., predictability analysis). 

However, it is also possible, if not even likely, that the saliency of different speaking styles may vary more in naturalistic environments. The average acoustic properties of naturalistic IDS may not reflect language input to children in those communicative scenarios or even certain time-points in utterances where attention control by stylistic means truly matters for learning, attentional control or affective regulation (see also \citeNP{MacDonald2020} for a discussion). When creating evaluation test sets for models under naturalistic settings, the above-mentioned aspects need to be considered for interpretation. Ultimately, the aim of the testing protocol also matters: the use of naturalistic data enables to test model behaviours ``in-the-wild'', and can lead to novel empirical hypotheses about the role and characteristics of IDS and ADS in real-life learning scenarios. However, this should be preceded by  \textit{model validation} on data for which comparable infant data on infant behaviour exists, which currently is largely limited to in-lab data from controlled experiments.

Concerning the audio quality, our results suggest that comparable and acceptable audio quality is reachable by manual annotation of the clips but also by using automatic tools for SNR estimation (at least for the acoustic features and modelling task investigated here). The practical implication is that automatic tools, such as that of \citeA{Brouhaha2022}, can be effectively used to screen out subsets of data that fit desired audio quality needs. This is especially useful when used in conjunction with other automated tools to process long-form datasets, as manual quality annotation does not scale up to hundreds or thousands of hours of speech.  

To sum up, the present study adds to the shared knowledge on how the recording conditions (in-lab versus at-home) play a role in analysis and computational modelling outcomes, in this case for IDS and ADS speech. Ideally, use of both in-lab and at-home data in tandem would enable a more comprehensive analysis of the phenomenon of interest. In addition, determining the audio quality of long-form recordings can be achieved with an automatic tool instead of tedious annotations, and with a low risk of missing the salient findings that are also present in data annotated as good quality by human listeners. In future work, a more comprehensive testing of the potential domain mismatch between model training and testing data is needed, for instance, by also training the models on naturalistic instead of clean data. However, this is left to future work.

\section{Acknowledgement}

Authors MACB and OR were supported by Academy of Finland grants no. 314602, 320053, and 345365, and Kone Foundation grant L-SCALE. AC was funded by Agence Nationale de la Recherche (ANR-16-DATA-0004 ACLEW), the ERC (ExELang, 101001095), and a J. S. McDonnell Foundation Understanding Human Cognition Scholar Award. We thank Marvin Lavechin and Maureen de Seyssel for their insights that benefited this research at the early stages. The measurements and scripts can be found on https://github.com/SPEECHCOG/ids\_audio\_quality\_analysis

\bibliographystyle{apacite}

\setlength{\bibleftmargin}{.125in}
\setlength{\bibindent}{-\bibleftmargin}
\bibliography{cogsci_template.bib}

\end{document}